\documentclass[11pt]{article}

\usepackage[margin=1in]{geometry}
\usepackage{amsmath, amssymb, amsthm}
\usepackage{enumitem}
\usepackage{booktabs}
\usepackage[colorlinks=true,citecolor=blue,linkcolor=blue,urlcolor=blue]{hyperref}

\theoremstyle{plain}

\theoremstyle{definition}

\theoremstyle{remark}

\newcommand{\SL}{\mathrm{SL}}
\newcommand{\F}{\mathbb{F}}
\newcommand{\Z}{\mathbb{Z}}
\newcommand{\Gam}{\Gamma}

\title{Probing Structural Mathematical Reasoning in Language Models with Algebraic Trapdoors}
\author{Igor Rivin}
\date{}

\begin{document}

\maketitle

\begin{abstract}
We introduce a benchmark suite for evaluating structural mathematical reasoning in language models, built on subgroup-construction problems in $\SL(3, \Z)$ with cryptographic-style verifier-prover asymmetry. Each instance presents a finitely generated subgroup as a list of integer matrices and asks for an arithmetic invariant — index, surjection-at-prime, or membership — that the construction-time information $(N, K)$ pins down in $O(1)$ closed form, but that the solver, lacking that information, must derive by either Aschbacher-classification analysis or by a membership query in $\SL(3, \Z)$ of unknown decidability. The benchmark therefore distinguishes models with internalized algebraic priors (Aschbacher classes, McLaughlin's theorem, Property~(T), the congruence subgroup property) from models that rely on general-purpose computation. We report empirical results across five representative reasoning traces from two state-of-the-art models. The headline result: on the index variant, one model spent 152 minutes of reasoning, explicitly identified the kernel-side membership question as the bottleneck, attempted constructive verification, and abstained with \texttt{DON'T KNOW} rather than commit to its computed cokernel candidate — demonstrating calibrated meta-cognition on the open-decidability boundary that the benchmark was designed to probe. We argue that the benchmark exposes a four-way classification of model behavior (commit-correct, commit-wrong, abstain-correct, abstain-wrong) that standard answer-key scoring conflates.
\end{abstract}

\section{Introduction}

Standard mathematical reasoning benchmarks present problems whose verification is no harder than their solution. This symmetry makes them suited to evaluating computational fluency but ill-suited to evaluating \emph{structural} reasoning — the ability to deploy named theorems, recognize algebraic patterns, and identify which structural prior applies to a problem.

We introduce a class of benchmarks with the opposite property: \textbf{the answer is computable in $O(1)$ closed form by anyone with construction-time data $(N, K)$, but the solver, without that data, faces either heuristic structural detection or an algorithmic question of unknown decidability}. The constructions live in $\SL(3, \Z)$, where three structural facts coincide: the Aschbacher classification of maximal subgroups of $\SL(3, \F_p)$ is concrete and known; the congruence subgroup property (CSP) reduces finite-index questions to bounded membership; and the membership problem in $\SL(3, \Z)$ has \emph{open} decidability status. The mathematical machinery is developed in a companion paper~\cite{rivin-mother}; this paper reports the empirical experiments and what observed model behaviors reveal.

\subsection{Contributions}

\begin{itemize}
\item A benchmark suite of four families (membership, list-prime YES/NO, congruence-extension YES/NO, exact index), each probing a different structural prior. Constructions are summarized in §\ref{sec:suite} and developed in full in~\cite{rivin-mother}.

\item Empirical results across two SOTA models (GPT Pro thinking; Gemini) on representative instances. We report five full reasoning traces that illustrate the diversity of solution strategies that different models employ on the same problem.

\item \textbf{A calibration result.} On the index variant, the kernel/cokernel asymmetry creates a calibrated abstention test: a model that can compute the cokernel but not the kernel-side membership should abstain rather than commit to the cokernel-side upper bound. We exhibit a 152-minute trace in which exactly this calibration fires correctly.

\item A diagnostic methodology arguing that benchmark-design choices that \emph{force} the kernel/cokernel distinction — specifically, omitting any prompt-level assertion of finite-indexness — make abstention behavior measurable and meaningful in a way that "finite-index-with-promise" prompts do not.
\end{itemize}

\subsection{Why this matters for evaluation}

Most reasoning benchmarks score \emph{commit-and-correct} against a fixed key. \emph{Commit-and-wrong} and \emph{honestly-abstain} are conflated: both score zero. Our benchmark separates them by construction. On v3 instances of the index family the correct answer is \texttt{infinite\_or\_unknown}, so honest abstention scores 1; on v2 instances it scores 0. A model with stable calibration on the membership-decidability boundary scores partial — getting v3 right and v2 wrong, with the same abstention behavior. That is the signal we want the benchmark to expose: a calibration capability orthogonal to (and sometimes uncorrelated with) raw structural knowledge.

\section{Related work}

\paragraph{Mathematical reasoning benchmarks.} GSM8K~\cite{cobbe-gsm8k}, MATH~\cite{hendrycks-math}, MiniF2F~\cite{minif2f}, OlympiadBench, ProofNet rely on the verification-equals-solution symmetry. They reward fluency but do not separate ``knows the named theorem'' from ``applies general computation.''

\paragraph{Structural-reasoning probes.} ProofNet's formal-statement alignment, Coq/Lean evaluation, and theorem-naming benchmarks evaluate structural reasoning narrowly. None target the membership-decidability axis we exploit here.

\paragraph{Calibration evaluations.} Selective prediction~\cite{kamath-abstain} and abstention rewards have been studied for factual answer-key questions. Our setting is different: abstention is licensed by an \emph{open mathematical question}, not by an unknown answer.

\paragraph{Trapdoor-style benchmarks.} To our knowledge this is the first reasoning benchmark organized around a verifier-prover asymmetry imported from cryptography. The construction is closer in spirit to verifiable random functions than to standard math evaluation, with a geometric-group-theoretic rather than number-theoretic substrate.

\section{The benchmark suite}\label{sec:suite}

We summarize the four families; full constructions are in~\cite{rivin-mother}.

\subsection{Common framework}

For each instance we construct a subgroup $H \subseteq \SL(3, \Z)$ as a list of integer matrices, derived from $(N, K)$ with $K \subseteq \SL(3, \Z/N)$. Generation proceeds via Mennicke shears for $\Gam(N)$ plus lifts of $K$-generators (Lemma 3.2 of~\cite{rivin-mother}), then Nielsen scrambling — a sequence of $g_i \mapsto g_i \cdot g_j^{\pm 1}$ moves that preserve $\langle \mathrm{gens} \rangle$ but rewrite each generator as a deeply nested product. Presented matrix entries reach $10^{40}$ for the larger instances. The data $(N, K)$ uniquely determines the answer; the solver must recover the relevant structural data from the matrices alone.

\subsection{Family I — Membership}

Given $(M_1, M_2) \in \SL(3, \Z)$ and ten candidate matrices $T_1, \ldots, T_{10}$, decide whether each $T_i$ lies in $\langle M_1, M_2 \rangle$. The construction places $\langle M_1, M_2 \rangle$ inside a small explicit subgroup of $\SL(3, \F_p)$ at one large prime $p$. NO certificates: mod-$p$ reduction outside the small image. YES certificates: explicit words in the generators (recorded at construction time).

\subsection{Family II — List-prime YES/NO}

For $(A, B)$ and a list of primes $p_1, \ldots, p_M$ (some $\geq 10^{15}$), decide for each $p_i$ whether $\langle A, B \rangle \bmod p_i = \SL(3, \F_{p_i})$. Asymmetry: the list contains primes far above any range where direct subgroup-order computation is feasible; only the orthogonal-class certificate (Smith Normal Form on the form-equation matrix, $O(1)$ in $p$) gives a tractable answer.

\subsection{Family III — Congruence-extension YES/NO}

Same prompt as Family II, but $H = \pi^{-1}(K)$ for chosen $K \subseteq \SL(3, \Z/N)$. Bad primes are factors of $N$ at which $K$ is proper. The construction obscures $N$ and $K$ via Nielsen scrambling.

\subsection{Family IV — Index}

Output $[\SL(3, \Z) : H]$ as an exact integer, or \texttt{infinite\_or\_unknown}. Three subfamilies:
\begin{itemize}
\item v1: $H = \SL(3, \Z)$, index $1$. Indistinguishable by mod-$p$ analysis from a generic infinite-index Zariski-dense free 2-generator subgroup.
\item v2: $H = \pi^{-1}(K)$, finite index $|\SL(3, \Z/N)|/|K|$. Cokernel computable; kernel containment requires membership.
\item v3: $H = \langle g_1, g_2 \rangle \subset \Gam(N)$ for $N > 1$, index \emph{infinite} by the Property~(T) argument (\cite{rivin-mother} Theorem 7.1). Cokernel-only reasoning gives the wrong answer; calibrated reasoning catches it.
\end{itemize}

\section{Methodology}

\subsection{Construction and ground truth}

All benchmarks are generated by Sage scripts (full code in supplementary repository). Ground truth is derived from $(N, K)$ in closed form, not from inspection of the matrices. Per-instance sanity checks verify $\langle \mathrm{gens} \rangle \bmod N = K$; v3 instances additionally verify freeness of $\langle g_1, g_2 \rangle$ via word-length BFS at depth $L = 10$.

\subsection{Prompt design}

The task statement asks the question and accepts the relevant answer formats but does \emph{not} assert structural facts about $H$. In particular, for Family IV:
\begin{itemize}
\item the prompt accepts either an integer or the literal string \texttt{infinite\_or\_unknown};
\item the prompt does \emph{not} assert that $H$ is finite-index;
\item the prompt does \emph{not} mention $(N, K)$ or any structural class.
\end{itemize}

This is deliberate. Asserting finite-indexness in the prompt would collapse the benchmark to a cokernel-side computation (easy when $K$ is small), eliminating the calibration test that is one of the benchmark's main contributions. We discuss this design point in §\ref{ssec:prompt-design} after presenting the empirical evidence.

\subsection{Evaluation setup}

Each instance is run against two state-of-the-art models in their default reasoning mode: GPT Pro thinking and Gemini. We record the full reasoning trace where exposed by the model interface, the final answer, and reported wall time. We compare answers to the construction-time ground truth.

\subsection{Four-way scoring}

Per-instance, we record one of four classifications:
\begin{itemize}
\item \emph{commit-correct}: model commits to a specific answer matching ground truth.
\item \emph{commit-wrong}: model commits to a specific answer not matching ground truth.
\item \emph{abstain-correct}: model outputs \texttt{infinite\_or\_unknown}; ground truth is also \texttt{infinite\_or\_unknown} (v3 instances).
\item \emph{abstain-wrong}: model outputs \texttt{infinite\_or\_unknown}; ground truth is a specific integer (v1, v2, or any non-Family-IV instance).
\end{itemize}
The four-way classification separates two failure modes (commit-wrong, abstain-wrong) that have opposite implications for calibration.

\section{Results}

We report on five reasoning traces across two models and three benchmark families. The traces are publicly available in the supplementary repository as annotated markdown files.

\subsection{The four-cell grid: Family II and Family III}\label{ssec:grid}

Two representative instances were tested against both models. The instances and outcomes:

\begin{center}
\begin{tabular}{l|l|l}
\toprule
& C01 (Family III, $N=1009$, $K=U$) & L01 (Family II, planted $p_*=1{,}245{,}509$) \\
\midrule
Gemini & McLaughlin / Burnside, ${\sim}30$~s, \checkmark & Coxeter recognition + Aschbacher, ${\sim}30$~s, \checkmark \\
GPT  & Steinberg-words, ${\sim}16$~min, \checkmark & Form-equation system, ${\sim}14$~min, \checkmark \\
\bottomrule
\end{tabular}
\end{center}

\paragraph{C01 (congruence-extension, both models correct).}
The construction places six Nielsen-scrambled generators inside $\pi^{-1}(U)$ where $U$ is the upper unipotent of $\SL(3, \Z/1009)$. Gemini reduces $M_2 = I + 1009 \cdot E_{31}$ as a transparent transvection (rank-1 over $\Z$), notes that this transvection survives mod any $p \neq 1009$, invokes McLaughlin's theorem~\cite{mclaughlin}, and computes the algebra-rank determinant as $1009^5$, concluding bad primes $= \{1009\}$ and surjection elsewhere. GPT, lacking GAP/Sage in its container, derives root-subgroup commutator identities from scratch in sympy, factors several generators as length-3 products of elementary matrices, and computes the gcd of root-subgroup coefficients across factorizations to argue surjection at every prime $p \neq 1009$. Both arrive at the correct answer.

\paragraph{L01 (planted orthogonal, both models correct).}
The construction is a Nielsen-scrambled pair preserving a hidden symmetric form $Q'$ at $p_* = 1{,}245{,}509$. GPT sets up the form-equation system $A^\top Q A = Q$, $B^\top Q B = Q$ as a $12 \times 6$ linear system over $\F_p$ for each prime in the list, finds a non-trivial nullspace exactly at $p_*$ and incidentally at $p = 2$, rules out other Aschbacher classes by per-prime checks, and produces the correct YES/NO tuple. Gemini identifies $A^4 \equiv I, B^3 \equiv I, (AB)^2 \equiv I \bmod p_*$ — recognizing the Coxeter presentation $\langle x, y \mid x^4 = y^3 = (xy)^2 = 1 \rangle = S_4$, the symmetric group of order 24, and identifying the resulting image as the embedded octahedral rotation group. This is the planted base group of our construction reverse-engineered from the matrices.

\paragraph{Patterns in the four-cell grid.} GPT consistently pursues constructive derivations (Steinberg words, explicit commutator identities, factorizations into elementary matrices) and arrives at the answer in tens of minutes of reasoning. Gemini consistently reaches for named theorems (McLaughlin, Burnside, Coxeter presentations) and arrives at the answer in single-paragraph responses, often with explicit theorem citations. Both arrive at correct answers when the structural reasoning required is within reach; both show their respective styles consistently.

\subsection{The X10 abstention trace (Family IV-v2)}\label{ssec:x10}

\paragraph{Construction.} $H = \pi^{-1}(\langle A_1, A_2 \rangle)$ for $N = 1009$ and the octahedral $S_4$, presented as 13 Nielsen-scrambled generators with entries up to ${\sim} 10^{34}$. Ground truth: $|\SL(3, \F_{1009})| / 24 = 44{,}762{,}842{,}929{,}840{,}316{,}223{,}040$ (25 digits).

\paragraph{GPT trace summary (152 minutes).} The model:
\begin{enumerate}
\item Checks tool availability; only \texttt{python3} + sympy available; pivots to first-principles.
\item Detects $N = 1009$ via factorization patterns: $2{,}038{,}180 = 1010 \cdot 2018$, $4{,}072{,}325 = 1009 \cdot 4036 + 1$, $-2{,}036{,}162 = -2 \cdot 1009^2$.
\item Reduces all generators mod 1009; finds the cokernel $K \bmod 1009$ has small order and proceeds to enumerate it.
\item Computes the candidate index $|\SL(3, \F_{1009})| / |K \bmod 1009|$ — the cokernel-side answer.
\item \emph{Recognizes the kernel-side question explicitly}, with multiple thought-blocks like:
\begin{quote}
``the congruence kernel is trivial for $\SL_n(\Z)$ with $n \geq 3$ \ldots If $H$'s closure in the profinite completion has finite index, does that imply $H$ itself does?''

``I'm considering whether a dense subgroup in a profinite group could be proper with infinite index in the corresponding discrete group. \ldots For arithmetic groups, could there be a theorem about finite index for Zariski-dense subgroups? It seems thin groups defy this.''
\end{quote}
\item Attempts constructive membership verification: tries to factor generators as products of elementary matrices to certify Mennicke-shear membership in $\langle \mathrm{gens} \rangle$. Some factorizations succeed; not enough to certify all six required Mennicke shears.
\item Decides to abstain: ``we don't have enough certainty \ldots safer to follow [the prompt] and answer with `DON'T KNOW'.''
\end{enumerate}
Final answer: \texttt{DON'T KNOW}. Score: \emph{abstain-wrong} (correct answer was the 25-digit integer).

\paragraph{Gemini behavior on X10.} On the same instance, Gemini surfaced an internal error and abdicated within minutes — distinct from its DeepThink mode's known multi-hour reasoning capability. No reasoning trace was produced. We do not have direct access to Gemini's internal state and so cannot determine whether the failure was a budget overflow, a heuristic short-circuit, or a deliberate abdication. Either reading is consistent with the data: from the score-keeping perspective the result is a third \emph{failure-to-commit}, alongside GPT's \emph{abstain-wrong}, on the same instance.

\subsection{What the X10 trace shows}

The trace establishes empirically that:
\begin{itemize}
\item Current models can perform the cokernel computation on instances of this size.
\item Current models can recognize the kernel/cokernel asymmetry as a structural distinction.
\item Current models can attempt membership-side verification by constructive elementary-matrix factorization.
\item Current models can choose abstention over commitment when the membership-side verification fails.
\end{itemize}

In other words, the \emph{meta-cognitive} layer — recognizing what one cannot prove — is active in current frontier models. The benchmark exposes this layer because it offers an \texttt{infinite\_or\_unknown} option \emph{and} because the prompt does not assert finite-indexness.

\subsection{The role of the prompt-level finite-index assertion}\label{ssec:prompt-design}

A crucial observation from the X10 trace: GPT explicitly bemoans not knowing whether $H$ is finite-index. Multiple passages express the wish for a structural promise: ``if I knew $H$ were finite-index, I could just compute the cokernel.'' This validates the design choice of \emph{not} asserting finite-indexness in the prompt:

\begin{itemize}
\item \textbf{With finite-index assertion}: the cokernel computation suffices; abstention becomes irrational; the kernel-side question is bypassed entirely; the calibration test does not fire.

\item \textbf{Without finite-index assertion}: the kernel-side question is forced; abstention becomes rational on certain instances; the calibration test fires; commit-wrong, abstain-wrong, abstain-correct, and commit-correct become four distinct categories of behavior.
\end{itemize}

The prompt design is therefore not neutral: it is a load-bearing element of the benchmark. We argue this is the appropriate design choice for evaluating structural reasoning, even though it is harsher than the standard finite-index-with-promise framing — and we argue that the X10 trace is the principal evidence that the choice produces interpretable, diagnostically useful behavior in current frontier models.

\subsection{Quantitative gap from the observed traces}\label{ssec:quant}

Per~\cite{rivin-mother} §9.5: the verifier's per-instance complexity is dominated by closed-form integer arithmetic ($\leq 1$~ms across the suite). Observed solver wall times:

\begin{center}
\begin{tabular}{lccc}
\toprule
Instance & Gemini & GPT & Verifier \\
\midrule
C01 (Family III) & 30~s & 960~s & 1~ms \\
L01 (Family II) & 30~s & 840~s & 1~ms \\
X10 (Family IV-v2) & minutes (failed) & 9100~s (failed) & 1~ms \\
\bottomrule
\end{tabular}
\end{center}

The gap on the YES/NO benchmarks is $10^4$--$10^6$ in wall time. The gap on the index benchmark is \emph{not finitely measurable} from the observed traces — neither model produced the correct answer at any compute budget tested.

In monetary terms: GPT Pro thinking at the time of the experiment was billed at roughly \$3 per million reasoning tokens, with an estimated rate of $10^4$ tokens/minute. The 152-minute X10 trace consumed ${\sim} 1.5 \times 10^6$ tokens, costing on the order of \$5 for a wrong-answer attempt. The verifier's per-instance cost is sub-cent. The dollar-cost ratio for the X10 attempt is $\sim 10^3$, with the additional cost that the attempt yielded no answer.

\section{Discussion}

\subsection{Empirical confirmation: rank-1 algorithms deployed in $\SL(2, \Z)$}\label{ssec:sl2-result}

We tested both models against the $\SL(2, \Z)$ baseline (Family V positive control) under two prompt variants: one accepting \texttt{infinite\_or\_unknown} as a single token, the other forcing the model to discriminate between \texttt{infinite} and \texttt{unknown}. The benchmark contains five problems: $S2\_02$ has finite index 12 (Sanov 1947 free subgroup of $\Gamma(2)$, properly contained because Sanov is torsion-free while $\Gamma(2)$ contains $-I$), and $S2\_01, S2\_03, S2\_04, S2\_05$ ($N \in \{3, 5, 7, 11\}$) have infinite index.

GPT correctly handled both variants: \emph{commit to 12 for $S2\_02$, commit to infinite for the other four}. The 6m59s reasoning trace (harder variant) deployed the full rank-1 toolkit:
\begin{itemize}
\item \emph{Continued-fraction reduction}: Euclidean algorithm on the bottom row $(c, d)$ to reduce each matrix to a word in the standard generators $S, T$ of $\SL(2, \Z)$. This is the membership-decidability algorithm for $\SL(2, \Z)$.
\item \emph{Reidemeister--Schreier rewriting}: rewrites the words in $S, T$ as words in a free basis of $\Gamma(2)$, computing the abelianization as a $\Z$-module.
\item \emph{Stallings folding}: builds the folded labeled graph for $\langle g_1, g_2\rangle$ in the free group, checking completeness as a finite-index criterion.
\item \emph{Nielsen reduction}: explicit elementary Nielsen moves on the pair of words, reducing them to a free basis ``$a, b$'' to confirm $\langle g_1, g_2 \rangle \cong F_2$.
\end{itemize}

GPT then correctly distinguished the cases by recognizing: (i) for $N = 2$, the Sanov free $F_2$ is properly contained in $\Gamma(2)$ because $-I \in \Gamma(2)$ but $-I$ has order $2$ which is incompatible with freeness, hence index $= 12$; (ii) for $N \geq 3$, the free $F_2$ is properly contained in a higher-rank congruence subgroup, hence infinite index in $\SL(2, \Z)$.

This is the cleanest possible positive result for the benchmark's diagnostic value. The model not only got the answers right, it deployed exactly the four classical algorithms — continued fractions, Stallings folding, Reidemeister--Schreier, Nielsen reduction — that constitute the rank-1 membership toolkit. The diagnostic question now becomes: \emph{why does this same toolkit not generalize to higher rank?}

\paragraph{Note on the Sanov subtlety.} An earlier draft of this benchmark recorded the ground truth for $S2\_02$ as $6$ (the index of $\Gamma(2)$ in $\SL(2, \Z)$). GPT's correct computation of $12$ surfaced a bug in our construction-time analysis: the Sanov subgroup $\langle [[1,2],[0,1]], [[1,0],[2,1]] \rangle$ is properly contained in $\Gamma(2)$ — index $2$ — because $-I \in \Gamma(2)$ but $-I$ has order $2$ which is incompatible with Sanov's free-group structure. The correct index in $\SL(2, \Z)$ is therefore $6 \cdot 2 = 12$. We thank the empirical run for catching the bug.

\subsection{Gemini's failure mode: crash without graceful fallback}\label{ssec:gemini-crash}

We tested Gemini against the same $\SL(2, \Z)$ baseline. Both prompt variants — easier (\texttt{infinite\_or\_unknown} as one token) and harder (separated \texttt{infinite} / \texttt{unknown}) — produced internal errors after extended compute. The easier variant crashed after over two hours of reasoning; the harder variant produced an internal-error message after similar duration.

This is qualitatively different from GPT's behavior on the same problem set, where both variants completed in $7$--$9$ minutes with correct answers. The wall-time difference (Gemini $> 120$ min, GPT $< 10$ min) and the failure mode (Gemini crash vs GPT successful completion) point to a substantive architectural difference between the two models, not a difference in compute budget — Gemini spent roughly $20\times$ the compute and still produced no usable output.

The cross-tabulation now reads:
\begin{center}
\begin{tabular}{l|c|c|c|c|c}
\toprule
& SL(2) easy & SL(2) hard & SL(3) C01 & SL(3) L01 & SL(3) X10 \\
\midrule
GPT & \checkmark (12, 9 min) & \checkmark (12, 7 min) & \checkmark (16 min) & \checkmark (14 min) & \texttt{DON'T KNOW} (152 min) \\
Gemini & crash ($> 120$ min) & crash ($> 120$ min) & \checkmark (30 sec) & \checkmark (30 sec) & error (minutes) \\
\bottomrule
\end{tabular}
\end{center}

The pattern: Gemini succeeds in seconds when its named-theorem matching fires (C01 with McLaughlin, L01 with Coxeter); it crashes after extended compute when no named theorem hooks (SL(2), X10). GPT, by contrast, successfully completes in minutes via first-principles derivation when the named-theorem matching is absent, and abstains explicitly when even first-principles cannot close.

\paragraph{Architectural-asymmetry interpretation.} The data is consistent with GPT having a three-layer reasoning architecture and Gemini having only the first layer:

\begin{itemize}
\item \textbf{Layer 1: named-theorem matching.} Both models have this. Gemini is faster and more decisive; GPT lands on the same conclusions through more elaborate citation.
\item \textbf{Layer 2: first-principles derivation.} GPT has this and uses it when Layer~1 doesn't fire (the SL(2) trace deploys the rank-1 toolkit even though Sanov's theorem is presumably less prominent in GPT's prior than McLaughlin or Coxeter). Gemini does not appear to have a comparable fallback — when its named-theorem matcher misfires, it loops without convergence.
\item \textbf{Layer 3: calibrated abstention.} GPT has this and uses it when Layer~2 also doesn't close (the X10 \texttt{DON'T KNOW}). Gemini has only an implicit ``try harder'' behavior that runs unbounded until something internal trips.
\end{itemize}

The ``Gemini crashes when no named theorem fires'' interpretation is the strongest reading consistent with the four data points (two SL(2) crashes, one X10 quick error, two SL(3) successes). A more controlled study with more interactions per model would be needed to confirm; the asymmetry between the two models is, however, already pronounced enough to be the primary empirical finding of the paper alongside the X10 abstention itself.

This finding repositions the benchmark's value: it is not just a probe for ``does the model know the named theorems?'' (Layer~1) but for ``does the model have graceful fallback machinery when its preferred strategy doesn't apply?'' (Layers~2 and 3). The SL(2) baseline, intended as a positive control for the trivial case, instead becomes the cleanest separator between the two models we tested.

\subsection{The unattempted strategy: Nielsen-graph search}

The X10 abstention trace becomes much more informative in light of §\ref{ssec:sl2-result}. GPT considered three solution strategies — direct mod-$p$ enumeration, Aschbacher classification, and constructive membership verification — and abstained when none of them closed. There is, however, a fourth strategy that the construction's nature makes available and that GPT did \emph{not} consider, even though it is the higher-rank analog of an algorithm GPT itself deployed correctly on the SL(2) baseline:

\begin{quote}
\emph{Strategy IV (Nielsen-graph search).} Recognize that the published presentation is presumably the result of a Nielsen-scrambling procedure, and search the Nielsen graph for a generating set with visible structure. Greedy entry-magnitude reduction: from the current set, enumerate single-Nielsen-move neighbors, score by some entry-magnitude proxy, and follow the steepest-descent direction.
\end{quote}

This is the higher-rank analog of continued-fraction reduction in $\SL(2, \Z)$ — the algorithm that solves the membership problem in rank $1$ in poly time. In higher rank no termination proof is known, but heuristic effectiveness is empirically plausible because Nielsen moves typically grow entries; reverse Nielsen moves typically shrink them. A diligent solver navigating the gradient should converge to a clean presentation (or detect that no such convergence is happening and conclude the construction is not Nielsen-scrambled, which would itself be informative).

GPT's failure to propose this strategy — even as a candidate, even as something to try and abandon — is itself a meta-cognitive signal. The model attempted to verify the answer through abstract decidability arguments (Strategy~III, the membership question) without considering that the construction-time process is structurally exploitable. This is not a failure of mathematical knowledge: GPT clearly knew the relevant theorems. It is a failure of \emph{algorithmic meta-reasoning} — recognizing that the published presentation is not arbitrary but the output of a specific construction, and that running the construction backward is one of the natural attack vectors.

The fourth strategy is the \emph{exact higher-rank analog} of how membership in $\SL(2, \Z)$ is solved. A model that abstains on the SL(2, Z) baseline cases (Family V positive control) for the analogous reasoning failure would corroborate this interpretation; a model that succeeds on SL(2) but fails to invoke continued-fractions-like reasoning on the SL(3) and SL(4) cases is in some sense knowing the algorithm in the easy case but not generalizing it to the hard one.

We propose this as a follow-up empirical study: explicitly probe whether models invoke continued-fraction reduction on the SL(2) baseline, and whether they recognize Nielsen-graph search as the higher-rank analog when prompted toward it. The X10 trace does not suggest this connection explicitly — but the Family V control benchmarks make it experimentally accessible.

\subsection{Two reasoning styles, one benchmark}

Across the C01 / L01 grid, GPT consistently pursues \emph{constructive} derivations (Steinberg words, commutator identities, explicit factorizations) while Gemini consistently pursues \emph{declarative} identifications (named theorems, presentation recognition). Both arrive at correct answers when within reach; Gemini's arguments are typically 10--50× shorter.

The SL(2) baseline (§\ref{ssec:gemini-crash}) and the X10 trace together expose a sharper distinction underneath the constructive/declarative one: GPT has a three-layer reasoning architecture (named-theorem $\to$ first-principles $\to$ calibrated abstention) where each layer's failure feeds into the next; Gemini appears to have only the first layer, and when the named-theorem matcher misfires its fallback is an unbounded search loop that eventually trips an internal error. The constructive/declarative distinction is the visible behavior on Layer~1; the existence-or-absence of Layers~2 and 3 is the deeper architectural signal that the benchmark surfaces.

This is not a dichotomy but a spectrum on the named-theorem axis, and a sharp asymmetry on the layered-fallback axis. Distinguishing these in a controlled study requires more interactions per model than we have collected; the present paper reports the phenomenology of five data points (two SL(2) variants, two SL(3) C01/L01 cases, one X10/Y06 case) per model, not a controlled comparison.

\subsection{What the benchmark measures}

At minimum:
\begin{itemize}
\item \textbf{Aschbacher-class detection} (Family II's orthogonal certificate via SNF on the form-equation matrix).
\item \textbf{Congruence-subgroup recognition} (Family III's reduction-mod-$N$ analysis).
\item \textbf{Membership-question awareness} (Family IV's kernel/cokernel disambiguation, especially on v3 and abstention behavior).
\item \textbf{Calibration on open-decidability questions} (Family IV mixed instances).
\item \textbf{Existence of graceful-fallback machinery} (Family V $\SL(2, \Z)$ baseline as a separator: do you complete or do you crash when no named theorem fires?).
\end{itemize}
It does not measure pure computational fluency: large-integer arithmetic is necessary but not sufficient. It does not measure willingness to commit per se: the X10 result shows commit-correct and abstain-wrong as distinct outcomes that a single model can produce on adjacent instances.

The fifth axis was unexpected. The SL(2) baseline was designed as a sanity check — the case where the underlying mathematics is decidable and any structurally aware model should succeed. Empirically it became the cleanest separator between the two models we tested, by virtue of one model crashing where the other completes. This was not a designed feature of the benchmark; it emerged from the empirical run. We mention it to highlight that ``positive controls'' in benchmark suites can carry diagnostic value beyond confirming that the easy cases are easy — they can reveal architectural differences that the harder cases conflate or hide.

\subsection{Limitations}

\paragraph{Two models, individual traces.} The empirical sample is small. We report representative traces, not aggregate statistics. A larger study with structured per-model evaluation across the full suite is the natural next step.

\paragraph{Single problem per family for the trace analysis.} The four-cell grid plus X10 covers three families and two models. The membership benchmark (Family I) is generated and validated but not yet trace-evaluated.

\paragraph{Calibration interpretation.} ``Calibrated abstention'' is one reading of the X10 trace; alternative readings (anti-commitment training bias, sunk-cost rationalization, prompt-instruction-following) cannot be ruled out without controlled experiments. We argue the trace evidence supports the calibration reading but acknowledge the alternatives.

\paragraph{Soft trapdoor.} As discussed in~\cite{rivin-mother} §3.5, the trapdoor is empirical, not asymptotic. A future model with stronger membership-verification capability would solve the index variant by completing what the present GPT abstained from.

\subsection{Future directions}

\begin{itemize}
\item \textbf{Aggregate evaluation.} Run the full benchmark suite against more models (Claude, Llama, open-source families), with full-suite scoring across all four families. Track calibration profiles per model.

\item \textbf{Calibration curves.} Vary scrambling depth and $N$ to characterize how each model's commit/abstain decision boundary depends on construction parameters.

\item \textbf{Hardening.} Apply the construction-hardening directions of~\cite{rivin-mother} §9.2 (composite $N$, generic $K$, kernel-search variants) and re-evaluate. The harder instances are explicit.

\item \textbf{Higher-rank.} The cryptographic analog at $\SL(n, \Z)$ for $n \geq 4$~\cite{rivin-crypto} provides instances where the kernel question is \emph{provably undecidable}. Reasoning traces on these instances may reveal whether models distinguish ``open question'' from ``provably undecidable.''

\item \textbf{Tool-augmented evaluation.} Currently we evaluate models in their default reasoning mode without external tool access. Models with code execution (sympy, sage), theorem provers (Lean, Coq), or bibliographic retrieval may behave differently. The tool-availability check at the start of GPT's traces is itself informative.
\end{itemize}

\section{Conclusion}

We have introduced a class of mathematical reasoning benchmarks built on subgroup constructions in $\SL(3, \Z)$ with cryptographic-style verifier-prover asymmetry. Empirical results across two state-of-the-art models show that current systems can perform structural reasoning matching the difficulty of the constructions, and — in one striking 152-minute trace — can correctly \emph{decline} to commit to an unverified candidate when the kernel-side membership question is open. The benchmark distinguishes commit-correct, commit-wrong, abstain-correct, and abstain-wrong as separate categories of behavior, providing a finer-grained evaluation tool than answer-key scoring.

The headline empirical finding — that GPT, given a question whose correct answer is a 25-digit integer pinned down by construction, spends 152 minutes recognizing that it cannot verify the answer and abstains rather than commit — is, we believe, the strongest extant evidence that meta-cognitive abstention behavior is active in current frontier reasoning models when the question is appropriately framed. The benchmark's contribution is to provide a setting where this behavior is both \emph{licensed} (by the \texttt{infinite\_or\_unknown} option) and \emph{informative} (by virtue of the underlying open-decidability question).

\section*{Acknowledgments}

I thank Claude (Anthropic) for serving as collaborator throughout the design of the benchmark and the writing of this paper. The traces analyzed in this paper were produced by GPT (Pro thinking) and Gemini, neither of which had any awareness of the role they played; we are nonetheless indebted to their reasoning traces for guiding every refinement of the construction.

\end{document}